\documentclass{article} 
\usepackage{collas2023_conference,times}
\usepackage{easyReview}
\usepackage{wrapfig}

\usepackage{amsmath,amsfonts,bm}

\def\eqref#1{equation~\ref{#1}}

\def\1{\bm{1}}

\DeclareMathAlphabet{\mathsfit}{\encodingdefault}{\sfdefault}{m}{sl}
\SetMathAlphabet{\mathsfit}{bold}{\encodingdefault}{\sfdefault}{bx}{n}

\usepackage{hyperref}
\hypersetup{
    colorlinks=true,
    linkcolor=red,
    filecolor=magenta,
    urlcolor=blue,
    citecolor=purple,
    pdftitle={Overleaf Example},
    pdfpagemode=FullScreen,
    }

\title{Active Class Selection for Few-Shot Class-Incremental Learning}

\author{Christopher McClurg \\
Pennsylvania State University\\
\texttt{cam7498@psu.edu}\\
\And
Ali Ayub \\
University of Waterloo\\
\texttt{a9ayub@uwaterloo.ca} \\
\And
Harsh Tyagi \\
Pennsylvania State University\\
\texttt{hkt5106@psu.edu}
\AND
Sarah M. Rajtmajer \\
Pennsylvania State University\\
\texttt{smr48@psu.edu}
\And
Alan R. Wagner\\
Pennsylvania State University\\
\texttt{alan.r.wagner@psu.edu} \\
}

\collasfinalcopy 

\begin{document}

\maketitle
\begin{abstract}
    For real-world applications, robots will need to continually learn in their environments through limited interactions with their users. Toward this, previous works in few-shot class incremental learning (FSCIL) and active class selection (ACS) have achieved promising results but were tested in constrained setups. Therefore, in this paper, we combine ideas from FSCIL and ACS to develop a novel framework that can allow an autonomous agent to continually learn new objects by asking its users to label only a few of the most informative objects in the environment. To this end, we build on a state-of-the-art (SOTA) FSCIL model and extend it with techniques from ACS literature. We term this model Few-shot Incremental Active class SeleCtiOn (FIASco). We further integrate a potential field-based navigation technique with our model to develop a complete framework that can allow an agent to process and reason on its sensory data through the FIASco model, navigate towards the most informative object in the environment, gather data about the object through its sensors and incrementally update the FIASco model. Experimental results on a simulated agent and a real robot show the significance of our approach for long-term real-world robotics applications. 
\end{abstract}
\section{Introduction}
\noindent  A primary challenge faced by robots deployed in the real world is continual adaptation to dynamic environments. Central to this challenge is object recognition \citep{ayub2020tell}, a task typically requiring labeled examples. In this work, we address the problem of parsimonious object labelling wherein a robot may request labels for a small number of objects about which it knows least. 

In recent years, several works have been directed toward the problem of Few-Shot Class Incremental Learning (FSCIL) \citep{tao2020fscil,ayub2020cognitively} to develop models of incremental object learning that can learn from limited training data for each object class. 
The literature has made significant progress toward developing robots that can continually learn new objects from limited training data while preserving knowledge of previous objects. However, existing methods make strong assumptions about the training data that are rarely true in the real world. For example, FSCIL assumes that in each increment the robot will receive a fully labeled image dataset for the object classes in that increment, and the robot will not 
receive more data for these classes again \citep{tao2020fscil,ayub2020cognitively,ayub2020tell}. 
In real world environments, however, robots will 
most likely encounter many unlabeled objects in their environment, and they will have to direct their learning toward a smaller subset of those unknown objects.

Active learning is a subfield of machine learning that focuses on improving the learning efficiency of models by selectively seeking labels from within a large unlabeled data pool \citep{settles2009active,ayub2022fewshot}. Related to active learning is active class selection (ACS) in which a model seeks labels for specific object classes \citep{lomasky2007active}. ACS can allow autonomous robots operating in real-world environments to focus their learning objects about which they know least. Most ACS models, however, have been designed for batch learning, i.e., they require all the previous training data to be available when learning in an increment~\citep{lomasky2007active}. Further, both active learning and ACS techniques have previously been tested on static datasets rather than with real agents/robots \citep{lomasky2007active,Yoo_2019_CVPR,Siddiqui_2020_CVPR}.

In this paper, we combine ideas from ACS and FSCIL to develop a framework that can allow an autonomous agent roaming in its environment to continually adapt by learning about the most informative objects through interaction with its human users. Toward this, we build on a state-of-the-art (SOTA) FSCIL model and extend it with techniques from ACS literature. We term this model Few-shot Incremental Active class SeleCtiOn (FIASco). We further integrate a potential field-based navigation technique with our model to develop a complete framework that can allow an agent to process and reason about its sensory data, navigate towards the most informative object in the environment, gather the data for the object through its sensors and incrementally update the FIASco model. We perform extensive evaluations of our approach in a simulated Minecraft environment and with a real robot in a laboratory setting. 
The main contributions of the paper are as follows: (1) We develop a novel framework extending FSCIL techniques with ideas from ACS and integrating it with autonomous agents. (2) Our experiments on a simulated and a real autonomous agent demonstrate the effectiveness and applicability of our framework for the long-term deployment of robots in real-world environments. Our code is available at \url{https://github.com/chrismcclurg/FSCIL-ACS}.

\begin{figure}[t]
\begin{center}
\includegraphics[width=0.9\columnwidth]{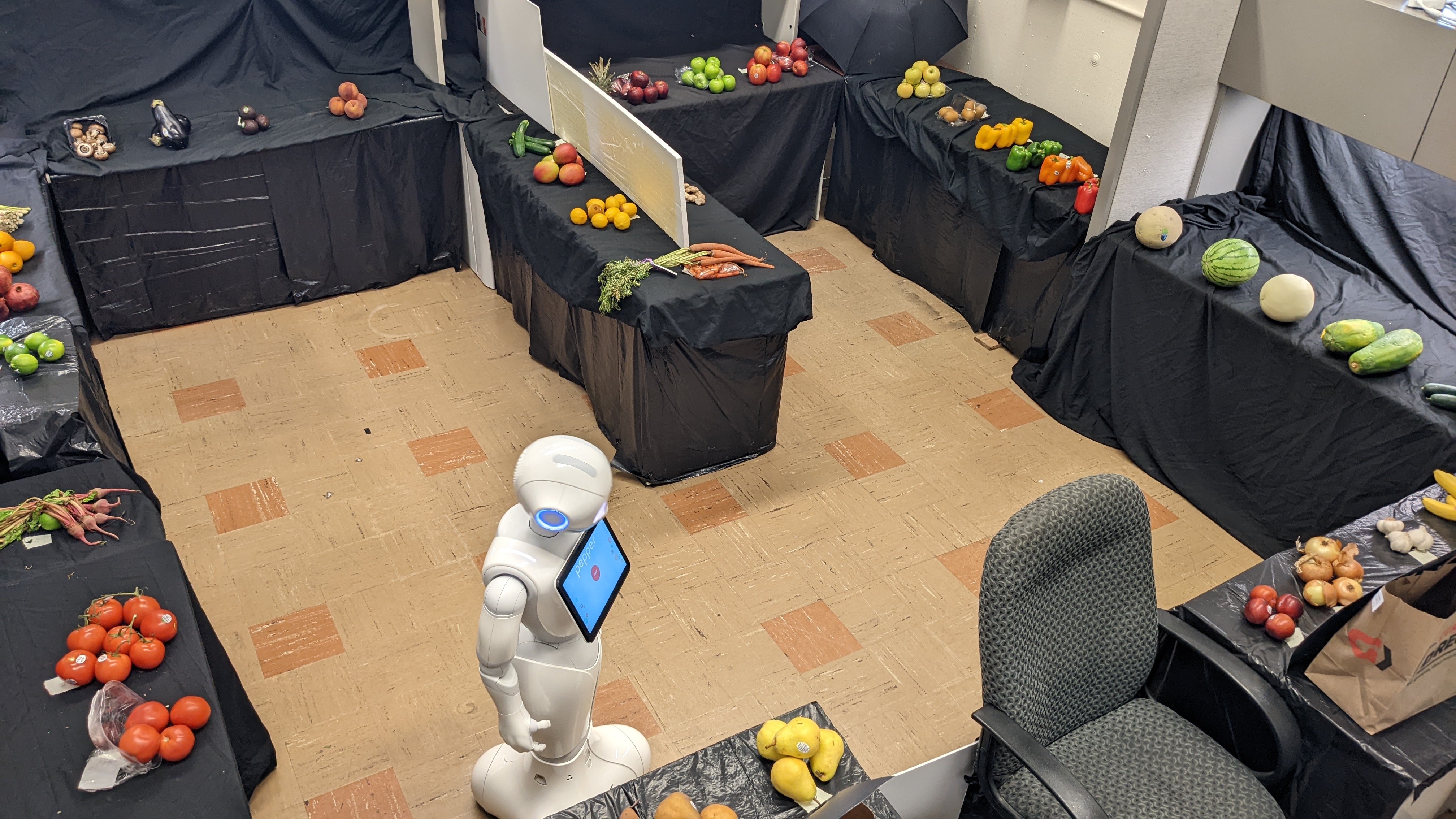}
\end{center}
\caption{Pepper learns about the environment by actively selecting classes to incrementally train on.} \label{top_pepper}
\end{figure}

\section{Background}
\label{background}
\noindent \textbf{Class-incremental learning (CIL)} considers the problem where labeled data is provided to the learner in increments of full classes. When applied to neural networks, CIL results in catastrophic forgetting, where the model forgets the previously learned classes and classification accuracy erodes~\citep{kirkpatrick2017overcoming}. A limitation of recent CIL methods is the reliance on storing a portion of the data from prior classes when learning new classes~\citep{Rebuffi_2017_CVPR,Castro_2018_ECCV,Wu_2019_CVPR}. These methods, often storing high-dimensional images, are not practical in situations when the system has limited memory. To avoid storing real images, some CIL methods use a regularization loss term to prevent the weights of the model from changing drastically when learning new classes ~\citep{kirkpatrick17,Li18,Dhar_2019_CVPR}. Other CIL methods regenerate images of the old classes with generative models~\citep{Ostapenko_2019_CVPR,ayub2021eec}. 

In a preliminary experiment, we compare the performance of a recent clustering approach (CBCL-PR)~\citep{ayub2020cognitively, Ayub_BMVC20, ayub2020cbclpr} against three popular CIL algorithms in a few-shot class-incremental learning setting: iCARL, PODNet, and DER. iCaRL~\citep{Rebuffi_2017_CVPR} stores exemplars in memory, uses a regularization term called distillation-loss \citep{Hinton15}, and Nearest Class Mean (NCM) to classify data~\citep{Mensink13,Dehghan19}. PODnet~\citep{douillard2020podnet} stores proxy vectors in memory, uses a spatially-based distillation-loss, and also uses an NCM classifier. DER~\citep{Yan_2021_CVPR} uses a two-stage approach that freezes previously learned representations and then augments the model with features from a fine-tuned extractor. The results of this preliminary experiment are contained in the appendix.

\noindent \textbf{Few-shot class-incremental learning (FSCIL)} adapts the class-incremental learning problem by limiting the number of training examples per class. Specifically, the data is first divided among training and test sets such that $x_i \in (X^{train} \cup X^{test})$, $y_i \in (y^{train} \cup y^{test})$. Then the training data is divided into increments $x_i^{train} \in (D_0^{train} \cup D_1^{train} \cup...D_n^{train})$, $y_i^{train} \in (C_0 \cup C_1 \cup...C_n)$ such that each increment is composed of a unique set of classes (i.e., $\forall{i,j} \ni i \neq j, C_i \cap C_j = \emptyset$). In the $i$-th increment, the model only trains on the corresponding training data $\{D_i^{train}, C_i\}$. The model is then evaluated on a test set that includes all classes seen so far (i.e., $\{\bigcup_{j=1}^{i}{D_j^{test}}, \bigcup_{j=1}^{i}{C_j}$). The size of an increment is $D_0^{train}$ containing $N_b$ of full classes. A problem setting which contains $N$ classes per increment and $k$ examples per class is known as $N$-way $k$-shot learning. In FSCIL, the problem is typically formatted with 100 full classes in the first increment, and then $10$-way $5$-shot learning for the remaining increments~\citep{tao2020fscil}. 

In a preliminary experiment, we compare the performance of CBCL-PR~\citep{ayub2020cognitively, Ayub_BMVC20, ayub2020cbclpr} against five other FSCIL algorithms: TOPIC, SPPR, Decoupled-DeepEMD, CEC, and FACT.  TOPIC~\citep{tao2020fscil} represents knowledge with a neural gas network in order to preserve the topology of the feature space. SPPR~\citep{zhu2021self} uses prototype learning, including random episode selection to adapt the feature representation and a dynamic relation projection between old and new classes. Decoupled-DeepEMD~\citep{Zhang_2020_CVPR} decouples the training of the embedding and the classifier; the embedding is trained on the initial increment of 100 full classes, while the subsequent increments replace class-specific classifiers with new mean embeddings. CEC~\citep{Zhang_2021_CVPR} trains an additional graph model to adapt prototypes of old and new classes. FACT~\citep{zhou2022forward} is the current state-of-art, which uses prototypes to limit the embedding space of old classes, reserving space for new classes. The results of this preliminary experiment are contained in the appendix.    
    
\noindent \textbf{Active Class Selection (ACS)} considers the problem where the learner can improve learning efficiency by requesting more data from a specific class~\citep{lomasky2007active}. In prior work, ACS was piloted to enable an artificial nose to efficiently learn to discriminate vapors~\citep{lomasky2007active}. In a batch learning setting, the learner used feedback from the previous batch to influence the class distribution among samples in the next class. A recent approach to ACS, PAL-ACS, demonstrated high performance by generating pseudo-examples, transforming an ACS problem into an active learning problem~\citep{kottke2021probabilistic}. This study was, however, limited to synthetic data.

\noindent \textbf{Active incremental learning} considers the problem where incremental learning and active learning are combined. In active learning, the learner may actively request labels for training data. One study assumed labels are no longer provided in the CIL setting~\citep{belouadah2020active}. Another study allowed a learner to incrementally select points for labeling from a point cloud~\citep{lin2020active}. A third study allowed a learner to incrementally select examples for annotation by a human expert ~\citep{brust2020active}. In these studies, the incremental learner selects training data to label, which defines the active learning problem. In contrast, this paper uses incremental learning to select classes to receive additional training instances, which is an active class selection problem.

\section{Model Description}

\begin{figure*}[h!]
\begin{center}
\includegraphics[width=\textwidth]{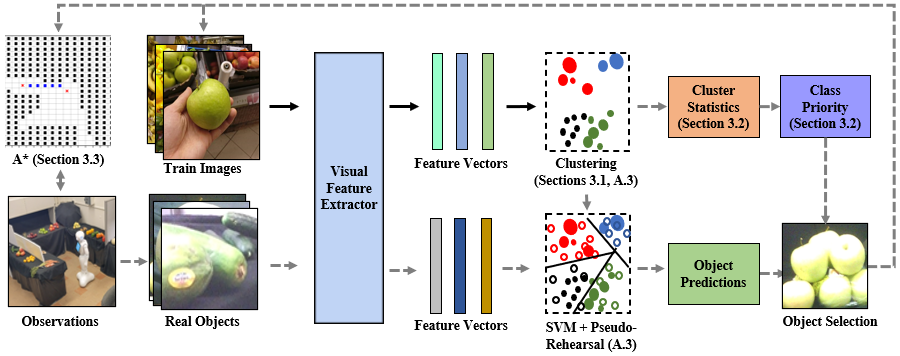}
\end{center}
\caption{This flow summarizes the training phase of FIASco. An agent uses a fixed feature extractor to obtain and cluster feature vectors from training images (solid line). The resulting centroids are used to fit a linear SVM, which is then used for predicting real objects. Cluster statistics are used to inform the agent which real objects to pursue and request more examples (training images). The training process combines few-shot class-incremental learning with active class selection.   \textit{*Please see the appendix for additional info on clustering or pseudo-rehearsal.} } \label{compModel}
\end{figure*}

\noindent Our goal is to develop a model (FIASco) that can not only learn incrementally, but can also select -- from observed classes in a novel environment-- classes which to receive more training instances. This problem is a modified class-incremental learning problem, whereas the next training class is determined by environmental availability and agent affinity. To learn incrementally, we ran preliminary experiments (see appendix) to identify CBCL-PR~\citep{ayub2020cbclpr,ayub_tcds} as the most promising approach for this problem. The identified approach not only produces SOTA results on few-shot incremental learning benchmarks, but also represents object classes as clusters, which have intrinsic statistics that can be used to to select the next training class in an environment. An overview of the model is shown in Figure~\ref{compModel}. In this section, we describe the components of FIASco, including incremental learning with clustering (Section \ref{Section:CBCLPR}), active class selection with cluster statistics (Section \ref{Section:ACS}), and navigation using a potential field created by cluster-averaged statistics of the observed classes in the environment (Section \ref{Section:Navigation}). 

\subsection{Incremental Learning with Clusters} \label{Section:CBCLPR}
\noindent In each increment, the learner receives the training examples (images) for new classes. Feature vectors of the images are generated using a pre-trained convolutional neural net as a feature extractor. Clusters are created from feature vectors that are within a tolerable distance of one another, enabling discrimination between classes and consolidation of these classes into long-term memory. For more details of this clustering approach, please see the appendix or related literature~\citep{ayub2020cognitively, Ayub_BMVC20, ayub2020cbclpr}. 

\subsection{Active Class Selection with Cluster Statistics} \label{Section:ACS}
\noindent We extend the learning approach to use feedback from cluster statistics. Specifically, the cluster space allows for measures -- cluster weight, class weight, and cluster variance -- to guide the selection of new samples for training. 

\begin{figure*}[h!]
\begin{center}
\includegraphics[width=\textwidth]{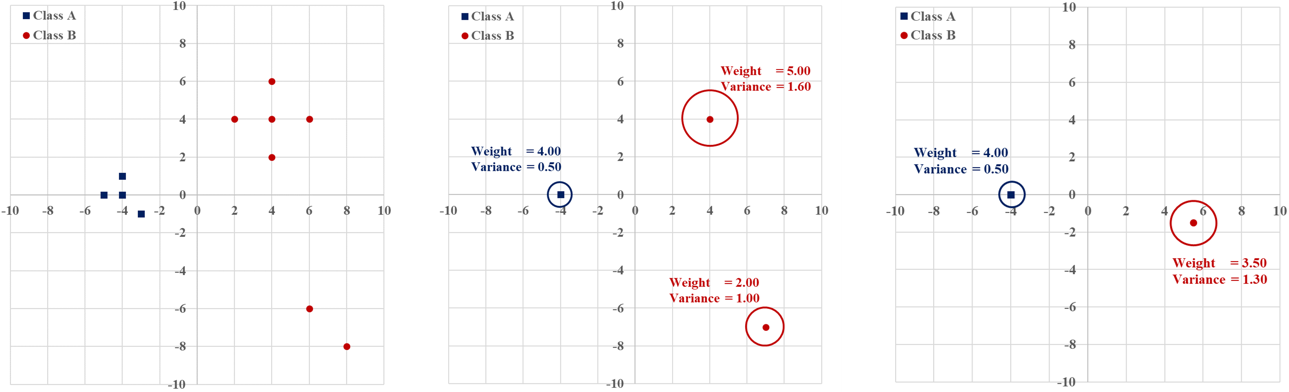}
\end{center}
\caption{\emph{Left}. An example distribution of data, where each point represents a training instance plotted in a two-dimensional feature space. \emph{Middle}. The clustering process groups similar training instances, extracting useful information such as cluster weight and cluster variance. \emph{Right}. The cluster-averaged class weight and class variance can be used for determining the next class to request. } \label{fig:metrics}
\end{figure*}

\noindent Cluster weight is the number of training examples included in an individual cluster within a class. Likewise, class weight is the number of training examples per class. Cluster variance is calculated in a recursive manner such that prior training data is not needed. As defined by Welford's method, the $n$-th update ($n>1$) of a cluster's variance is $s_n^2$ ~\citep{welford1962note,knuth2014art}: 

\begin{equation}
\begin{aligned}
(n-1) s_n^2 - (n-2) s_{n-1}^2 = (x_n-\bar{x}_n)(x_n-\bar{x}_{n-1})
\label{eq:variance}
\end{aligned}
\end{equation}

These internal measures give direct feedback for active class selection (ACS). Recall that previous ACS methods use results from the previous batch as feedback to specify the distribution of classes in the next batch. In incremental learning, the learner does not control the size of new batches. Therefore, class selection is instead an ordering of preferred classes:  

\begin{enumerate}
    \item \emph{Low Class Weight:} Prioritize classes with lower class weight. The intuition for this ordering is that adding instances to a class with fewer instances will likely add useful information (new clusters), increasing overall accuracy.
    
    \item \emph{Low Cluster Weight:} Prioritize classes with lower average cluster weight. The intuition for this ordering is that adding instances to classes with undeveloped clusters (outliers) will be more likely to impact (shift/ add weight to) the class-specific space, increasing overall accuracy. 

    \item \emph{Low Cluster Variance:} Prioritize classes with lower average cluster variance. The intuition for this ordering is that adding instances from classes with less noise will likely add valuable information with minimal overall noise. 
    
    \item \emph{High Cluster Variance:} Prioritize classes with higher average cluster variance. The intuition for this ordering is that adding instances from classes with more uncertainty will likely provide more distinct clusters within the class.  
    \end{enumerate}

To further illustrate these measures, consider a distribution of two classes of data, as shown in Figure~\ref{fig:metrics}. Each instance of data is initially plotted in the two-dimensional vector space (left). The clustering process (middle) extracts useful cluster information, such as weight and variance. Finally, the extracted information can be cluster-averaged per each class of data (right). Which class of data should be requested next for the purpose of training? According to the low class weight metric, class A should be requested $(4.0 < 7.0)$. According to the low cluster weight metric, class B should be requested $(3.5 < 4.0)$. For low cluster variance, class A should be requested $(0.5 < 1.3)$. Of course, class B should be requested for high cluster variance $(1.3 > 0.5)$. 

\subsection{Navigation from Active Class Selection}\label{Section:Navigation}

\begin{figure*}[h!]
\begin{center}
\includegraphics[width=0.7\textwidth]{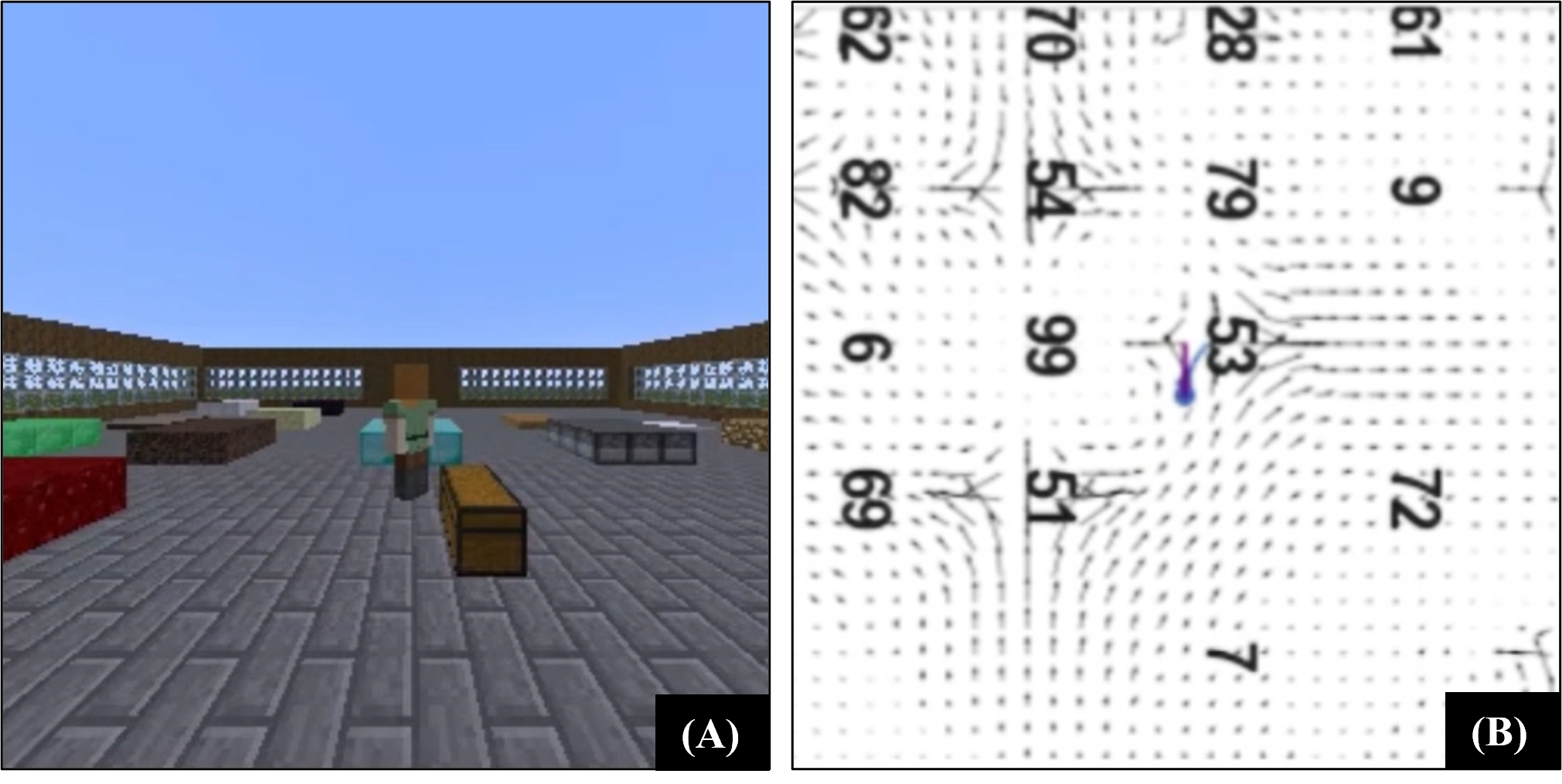}
\caption{A view of the agent in simulation (A) and the potential field created for navigation (B). } \label{fig:nav0}
\end{center}
\end{figure*}

\noindent Integrating our incremental ACS approach on an autonomous agent requires developing a method for navigation to move towards the most informative data samples. The selected method for navigation was a potential field approach, simplified from~\citep{koren1991potential}. Figure~\ref{fig:nav0} shows a potential field created from agent observations in the simulation.

\noindent Motivated to apply these methods on a real robot that can make some inference about distal objects ($d \leq d_{far}$) and then identify objects at a closer distance ($d \leq d_{close} < d_{far}$), the learner is given similar characteristics. In experiment, distances for class identification and feature extraction were set to $d_{far}$ and $d_{close}$, respectively. Objects within distance $d_{far}$ would be included in the learner's internal potential field, where the true class label would be known by the robot (i.e., close enough to ask a person for the true labels). For the $i$-th object in the potential field, an attractive or repulsive force $f_i$ was assigned based on the order of class priority determined in ACS. The potential field is then defined by equation~(\ref{potField}), where the $i$-th observation is made at ($x_i$, $y_i$) and the robot position is ($x_0$, $y_0$). Objects are only learned when the robot is within the distance $d_{close}$, where an image can be taken and features extracted for training. 

\begin{equation}
(F_{x}, F_{y}) = \Bigg(\sum_{i=1}^{n_i} \frac{f_i}{x_i - x_0}, \sum_{i=1}^{n_i} \frac{f_i}{y_i - y_0}\Bigg)
\label{potField}
\end{equation}

A common problem with potential fields is that the agent can get stuck in a local minima. Past solutions for this local minima problem have included adding small, random perturbations or adjusting the gain of a particular contribution to a potential field~\citep{arkin1989motor}. In our simulated experiment,  the number of time steps spent inside a relative location is counted. If the learner exceeds a specified count limit, it is directed back to the start position. Every time the learner returned to the start position, it is sent in a new direction (i.e., if the learner came from the North, it is randomly sent East, South, or West). 

\begin{figure*}[h!]
\begin{center}
\includegraphics[width=0.7\textwidth]{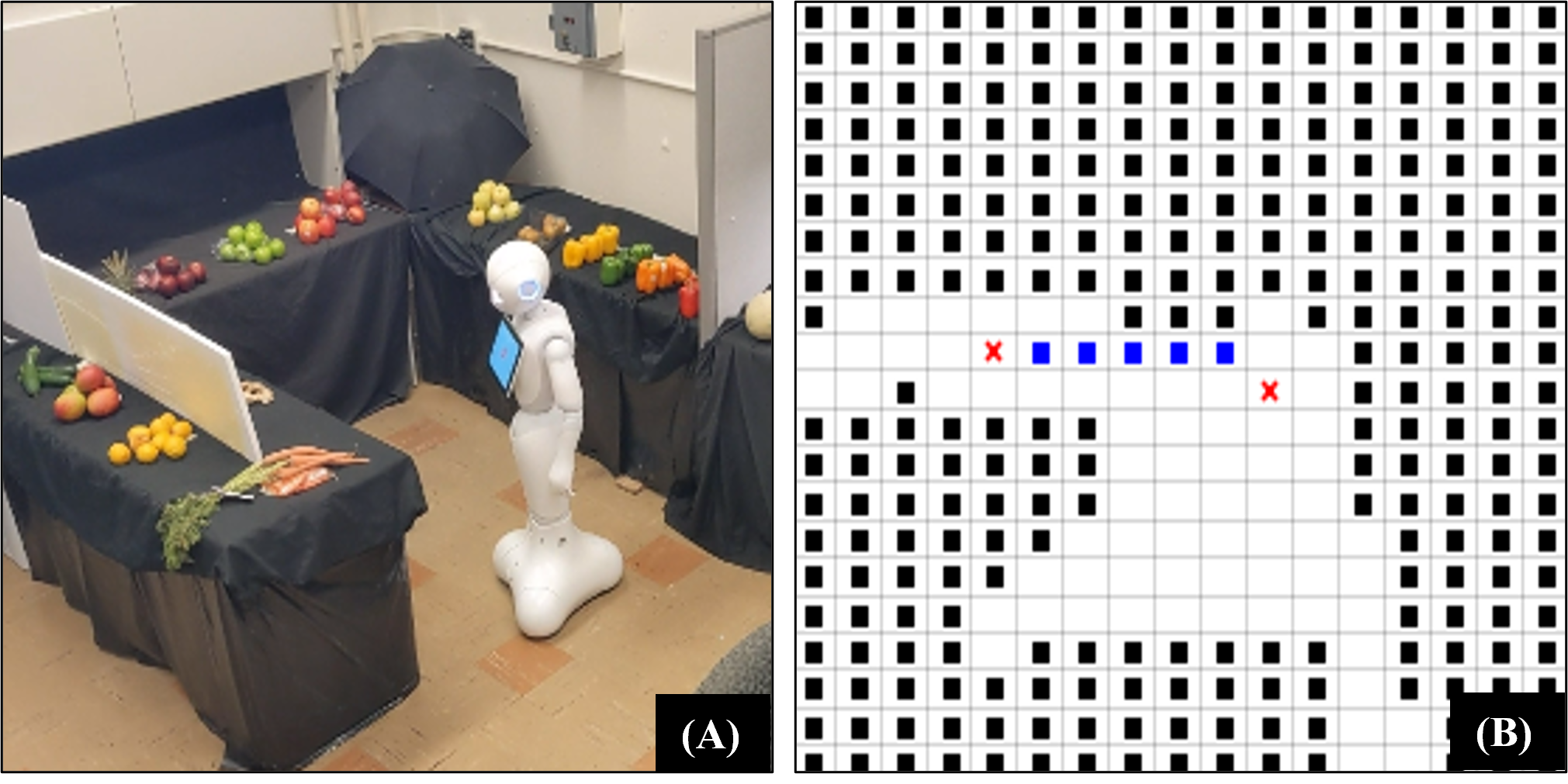}
\caption{A view of the robot in the environment (A) and the A* path created for navigation (B).} \label{fig:nav1}
\end{center}
\end{figure*}

In our experiment with a real robot, sensor error also presented problems. That is, not only is there possibility of getting stuck in a local minima, an undetected obstacle would also prevent movement of the robot. To mitigate the effects of sensing error, rather that use a continually-adapting potential field, the robot observed its surroundings once, then used A* path-planning~\citep{hart1968formal} to get to the location of selected class. This navigation method has less benefit for actively selecting classes; please see the results from Section~\ref{resultsDiscussion} for a discussion, or the appendix for more info on the A* method. Figure~\ref{fig:nav1} shows the A* path planning from robot observations in the environment. 

\section{Experiment: FSCIL-ACS in Minecraft} \label{Section:exp_malmo}
\noindent Our first experiment is an image classification task within the Minecraft simulation environment. We aim to show that a simulated robot can use internal feedback based on what it has learned about the environment (cluster space) to more efficiently seek unknown objects in the environment.

\subsection{Experimental Setup}
\noindent\textbf{Overview. }A robot in Minecraft is given two minute intervals to search the environment for new visual examples of objects. The robot navigates with an internal potential field, created from objects within an observable distance ($d < d_{far} = 15$). The robot can observe visual examples of an object \emph{only} when it stands over that object ($d < d_{close} = 1$). After the interval of searching, the robot processes the visual examples by updating its cluster space (FIASco) or re-training on all of the previous training data (SVM). Finally, the robot makes predictions on the test data (static subset of original dataset) and classification accuracy is recorded. The robot's affinity to different classes of items is updated using the ACS methods described in Section~\ref{Section:ACS} , which directly affects the future potential field for navigation. The experiment continues for 360 minutes. Please see the supplemental material for experiment replication notes.

\begin{figure*}[h!]
\begin{center}
\includegraphics[width=0.95\textwidth]{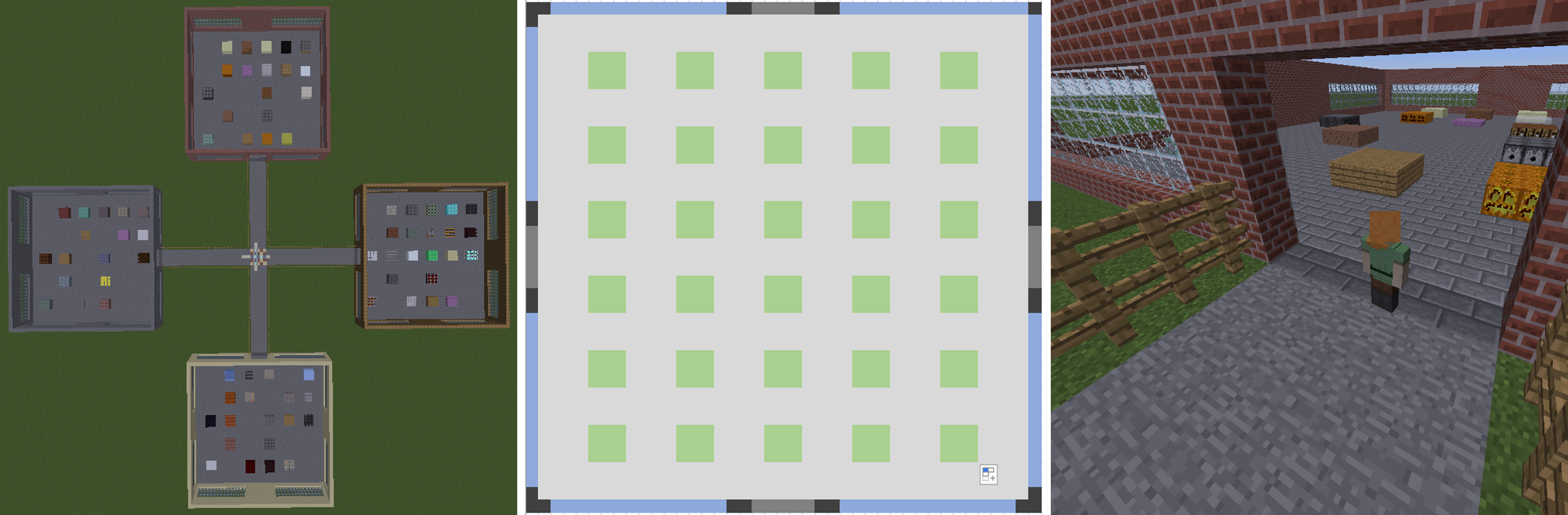}
\end{center}
\caption{\emph{Left}. The map depicts the layout of the simulation environment. \emph{Middle}. A single building has 30 containers of items, which are arranged randomly. \emph{Right}. The agent navigates between buildings, looking for particular items.  } \label{map}
\end{figure*}

\noindent\textbf{Baselines.} Cluster-based ACS methods were compared with a batch learner using `uniform' and `redistricting' class selection. The `uniform' method randomly sets the class order so that all classes have an equal opportunity to be prioritized. The `redistricting' method uses cross-validation to determine the most volatile (changing predictions when new samples are added in the validation stage) classes to prioritize. The cluster-based ACS methods are described in Section~\ref{Section:ACS}. Note that `uniform' is also run for FIASco and that `high cluster variance' is most similar to the previous `redistricting' method without the time-consuming validation step.  

\noindent\textbf{Environment. } Minecraft was used because it offers a large number of items and user control to create maps, enabling a realistic, yet constrained spatio-temporal situation for an agent~\citep{johnson2016malmo}. The experiment map (Figure~\ref{map}, left) contained four buildings. These buildings housed four unique groups of classes, grouped by the similarity of class-averaged feature vectors (centroids). Within a building, items were randomly, uniquely assigned to one of the thirty containers (Figure~\ref{map}, middle). These containers served as the link to real-world items. As an agent approached the location of a container, it would observe a certain type of Minecraft item. This observation was then mapped to a class of the training dataset. While in the proximity of a container, the agent could choose to learn about the class by standing directly over the container. In this case, the agent would receive a random 5-9 instances of a class for training, after which the container would be empty. The container does not restock until the next round of exploration, after the agent trains and updates its class affinity. 

\noindent\textbf{Data.} Two datasets were used for training and testing of the image classifier: CIFAR-100 ~\citep{krizhevsky2009learning} and the Grocery Store~\citep{klasson2019hierarchical} datasets. CIFAR-100 contains 60,000 32x32 images, evenly distributed among 100 classes. The classes include various types of objects, such as ``beaver'' or ``rocket.'' The Grocery Store dataset contains 5,125 348x348 pixel images, non-uniformly distributed among 81 classes. The classes include various goods found in grocery stores, such as types of fruits, vegetables, and packages. Both datasets were modified to have a 90:10 stratified train-test split. Please see the Appendix for more information about the data selection.

\begin{figure*}[h!]
\begin{center}
\includegraphics[width=\textwidth]{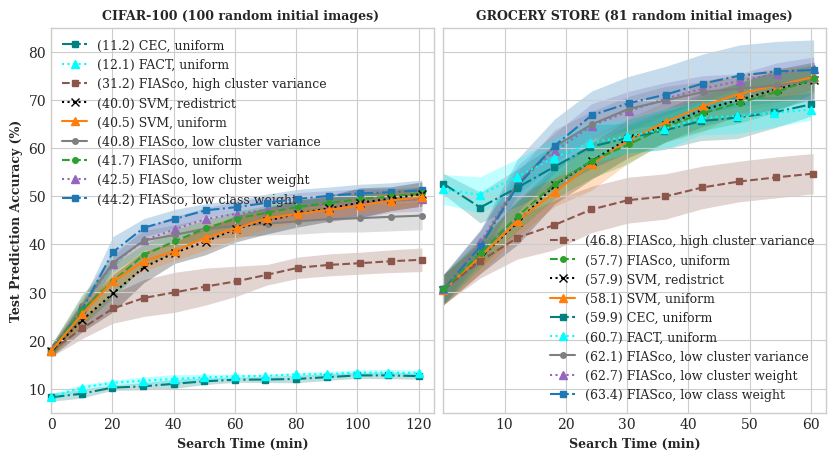}
\end{center}

\caption{Test prediction accuracy over time in Minecraft simulation. Note that SVM classifier is a batch learner, while FIASco, CEC, and FACT do not re-use training data. Average incremental accuracy is indicated.} \label{plot_malmo}
\end{figure*}

\noindent\textbf{Implementation. } The fixed feature extractor in this experiment was a Resnet-34 model pre-trained with Imagenet. The test was run with ten random seeds and the average was determined. For clustering, the distance threshold $D$ and number of pseudo-exemplars $N_{P}$ were determined by validation. For the CIFAR-100 test, the values for $D$ and $N_{P}$ were set to $17$ and $5$, respectively. For the Grocery Store test, the values for $D$ and $N_{P}$ were $15$ and $40$, respectively. For batch learning, a support vector machine with a linear kernel was used~\citep{boser1992training} to make test predictions given all extracted features.

\subsection{Experimental Results }
\noindent Results are shown in Figure~\ref{plot_malmo}. The metric used for comparison was average incremental accuracy. Note that the accuracy computed in this experiment is different from the preliminary study: rather than testing over only \emph{seen} classes, the learner is tested over \emph{all} classes in the environment. The highest performer in the CIFAR-100 test was FIASco with `low class weight' ACS ($44.2\%$), an improvement of $3.7\%$ over the best case of batch learning `uniform' ACS. The highest performer in the Grocery Store test was FIASco using `low class weight' ACS ($63.4\%$), an improvement of $5.3\%$ over the best case of batch learning `uniform' ACS. 


\section{Experiment: FSCIL-ACS with Pepper} \label{Section:exp_pepper}
\noindent In the final experiment, a Softbank Pepper robot was tasked with an image classification in an indoor environment. We aim to demonstrate that a real robot can use active class selection to more efficiently seek unknown objects (see Figure~\ref{top_pepper}).

\subsection{Experimental Setup} \label{setup2}

\noindent\textbf{Overview. }The robot is given sixty iterations to search the environment for new visual examples of objects. An iteration consists of the robot (1) relocating, (2) searching, (3) choosing an object, and (4) receiving training examples. To relocate, the robot first rotates with range sensors to define a localized map; an end location is chosen among the free space, and A* path planning is used~\citep{hart1968formal}. To search, the robot uses a top camera, which provides up to 2560x1080 pixel resolution at 5 fps. After taking images of the surrounding area, the robot uses the YOLO algorithm~\citep{redmon2016you} pre-trained on the Microsoft COCO dataset~\citep{lin2014microsoft} for object localization. To choose an object, the robot uses centroids for (initially weaker) classification with active class selection to pick the most desirable class. To receive training examples, the robot shows the human experimenter an image of the desired class, for which the human can give the true label of the predicted class, as well as ten visual examples. After every iteration, the robot updates its cluster space of learned classes. The robot's affinity to different classes of items is updated using the ACS methods. At the end of every three iterations, the robot makes predictions on the test data and classification accuracy is recorded. 

\noindent\textbf{Baselines. }Cluster-based ACS methods (Section~\ref{Section:ACS}) were compared with a batch learner using `uniform' class selection, which randomly sets the class order so that all classes have an equal opportunity to be prioritized. 

\noindent\textbf{Environment. } This test was completed in an indoor environment, where items were purchased from a local grocery store to represent classes in the Grocery Store dataset~\citep{klasson2019hierarchical}. Black cloths were used to cover tables and serve as a backdrop for items. Please see supplemental materials for images of the included classes. 

\noindent\textbf{Data.} The Grocery Store dataset~\citep{klasson2019hierarchical} was used for training and testing of the image classifier, as in Section~\ref{Section:exp_malmo}. The continually-trained image classifier was used for object recognition of the real objects in the experiment. A subset of 41 classes of the Grocery store dataset was used, comprised of items that could be primarily stored at room temperature. The dataset was modified to have a 90:10 stratified train-test split. Real items were distributed randomly by their coarse labels, such that similar items were grouped together (e.g., Red Delicious and Yellow Delicious apples). Please see the Appendix for more information about the data selection.

\noindent\textbf{Implementation. }The fixed feature extractor in this experiment was a Resnet-34 model pre-trained with Imagenet. For clustering, the distance threshold $D$ and number of pseudo-exemplars $N_{P}$ were determined by validation. For this test, the values for $D$ and $N_{P}$ were $15$ and $40$, respectively. For batch learning, a support vector machine with a linear kernel was used~\citep{boser1992training} to make test predictions given all extracted features.

\subsection{Experimental Results }\label{resultsDiscussion}
\begin{wrapfigure}[26]{R}{0.5\textwidth}
    \begin{center}
    \includegraphics[width=0.45\textwidth]{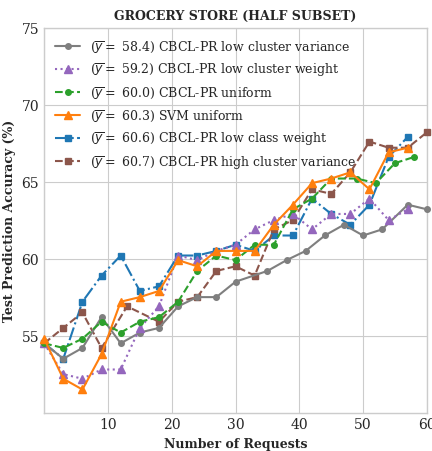}
    \end{center}
    \caption{Test prediction accuracy over iterations in indoor environment with Pepper. Note that SVM classifier is a batch learner, while FIASco does not re-use training data.} 
    \label{plot_pepepr}
\end{wrapfigure}

\noindent Results are shown in Figure~\ref{plot_pepepr}. 

The metric used for comparison was average incremental accuracy. The accuracy computed in this experiment is the same as in Section~\ref{Section:exp_malmo}: the learner is tested over \emph{all} classes in the environment. The highest performer in the test was FIASco with `high cluster variance' ACS ($60.7\%$), an improvement of $0.4\%$ over the best case of batch learning `uniform' ACS. 

While both experiments have a learner using the same measures to prioritize classes, there is a difference in the value of particular measures (e.g., high cluster variance.) This difference is likely due to the slight change in process, where the robot learner is making an initially weak prediction about the detected object classes before requesting a class (see Section~\ref{setup2}). Hence, a wrong prediction about a class with a high variance may actually provide valuable insight into the divisions of nearby classes. 

In terms of average incremental accuracy, the FIASco model does not show as much improvement over ACS with SVM, as compared to the simulated experiment. This result is likely due to the limitations in real navigation, noted in Section~\ref{Section:Navigation}. When the agent moved in simulation, the potential field was updated at every time step, calculating attractive weights for each new position of observed class. In the real environment, the robot made one full turn to observe its surroundings, then followed a path prescribed by A*. As the robot moved, new class observations were not included as options for the robot. The reason for this change was due to our particular robot being susceptible to drift error and sensor noise; we choose to reduce the sensing demand such that the robot would not get itself stuck as frequently. Note that the navigation method is kept constant in each experiment, so the comparison of ACS methods still holds true. In future studies, it would be helpful to improve the robot controller so that the more reactive navigation method could be used. 


\section{Conclusion}
\noindent To the authors’ knowledge, active class selection (ACS) has not previously been combined with few-shot incremental learning (FSCIL). This paper extends an incremental learner to use cluster statistics as feedback for actively selecting classes to learn. We have shown that the selected incremental learner (CBCL-PR) is not only state-of-the-art in a pure few-shot class incremental learning setting, but also that the cluster space is valuable to intrinsically motivate the learner to select specific classes. In both Minecraft simulation and real indoor environments, a robot that used cluster statistics for active class selection out-performed uniform batch-learning. 

A challenge of any (machine) learner is to gather labeled data for supervised training. We lay the groundwork for more efficient gathering and usage of labeled data, relaxing previous assumptions that have hindered the feasibility of robot learning. As opposed to previous methods in FSCIL, we do not rely on a prescribed class order, nor require training on half the dataset prior to incremental learning. These assumptions are both unrealistic and not applicable to a robot learning in a new environment. As opposed to previous methods in ACS, we incorporate more current efforts of incremental learning such that computational complexity is more favorable in the long-term (see Appendix).  

Future work should build on the merging of active class selection and incremental learning. The most obvious reason is that it is critical to bridge the gap between robot and agent learning. Additionally, there is opportunity to further advance the state-of-art in FSCIL-ACS. For instance, in the context of clustering, a combination of statistics could be used to guide class selection. More broadly, alternative internal measures could be used as feedback for class selection. Regardless, the advantages in combining ACS and FSCIL motivate a new direction for robot learning.


\subsubsection*{Acknowledgments}
This material is based upon work supported by the Air Force Office of Scientific Research under award number FA9550-21-1-0197.


\bibliography{main}
\bibliographystyle{collas2023_conference}


\appendix
\section{Appendix}
\subsection{Preliminary Study}
\noindent\textbf{Overview. }Two settings for few-shot class-incremental learning are considered, denoted \emph{traditional} and \emph{pure} FSCIL. In \emph{traditional} FSCIL, as described in Section~\ref{background}, the learner receives $N_b$-base classes of full data in the first session $(t = 1)$, then incrementally learns on the remaining data using $N$-classes per session with $k$-examples per class (i.e., $N$-way, $k$-shot) on subsequent sessions $(t > 1)$. The primary difficulties for learning in this setting include catastrophic forgetting and over-fitting due to class imbalance. In \emph{pure} FSCIL, as introduced here, the learner receives no base classes of full data, but rather incrementally trains with $N$-way $k$-shot learning from the first session $(t \geq 1)$. The primary challenges of this setting include catastrophic forgetting and learning without a large portion of the dataset. 

\noindent\textbf{Baselines. }CBCL-PR was compared with nine other methods: CBCL, iCarL, PODNet, DER, TOPIC, SPPR, Decoupled-DeepEMD, CEC, and FACT. The methods formulated for class-incremental learning (iCarL, PodNet, and DER) were adapted from previous works~\citep{zhou2021pycil,douillard2020podnet} to limit the number of shots. FACT and CEC were re-run from original code for the \emph{pure} FSCIL setting. Note that TOPIC and Decoupled-DeepEMD did not have full code available for reproduction, while SPPR performed significantly worse with any reduction in $N_b$ base classes, so these methods are excluded from \emph{pure} FSCIL comparison.

\noindent\textbf{Data.} The Caltech-UCSD Birds 200 (CUB-200) image dataset was used for a preliminary study~\citep{399}. CUB-200 contains 11,788 images, uniformly distributed over 200 classes. The classes are different species of birds.

\noindent\textbf{Implementation. }All methods used Resnet-18 pre-trained with Imagenet as a backbone. The preliminary study used ten random seeds. As in previous work~\citep{tao2020fscil}, \emph{traditional} FSCIL incorporated 100 base classes and 10-way 5-shot incremental learning. The first session $(t = 1)$ trained for 50 epochs with an initial learning rate of 0.1, decreased to 0.01 and 0.001 at 30 and 40 epochs, respectively. Subsequent sessions $(t > 1)$ were trained with a learning rate of 0.01 for 100 epochs. The mini-batch size was 128. For \emph{pure} FSCIL, no base classes were used, allowing 10-way 5-shot incremental learning from the first session. The learning rate was 0.01 for 100 epochs for all sessions $(t \geq 1)$. The mini-batch size was 64. 

\begin{figure*}[h!]
\includegraphics[width=\textwidth]{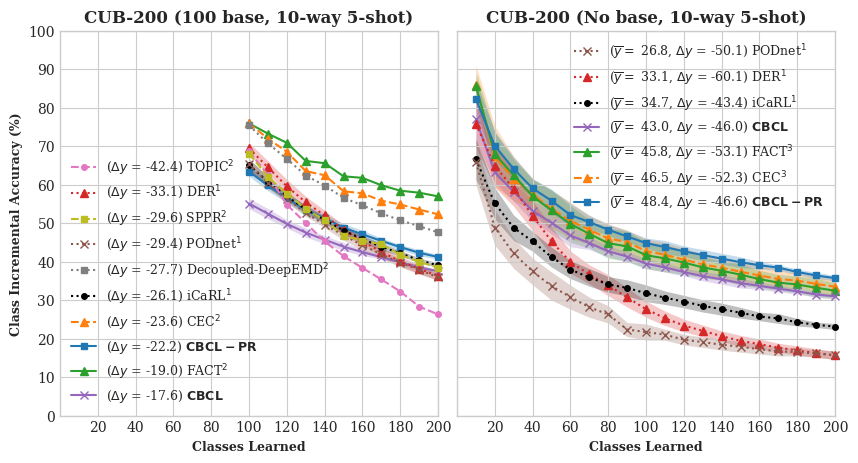}
\caption{A comparison of \emph{traditional} and \emph{pure} FSCIL on the CUB200 dataset. Note that $^1$CIL is modified for FSCIL, $^2$results are as reported, and $^3$traditional FSCIL is modified for pure FSCIL. Average incremental accuracy ($\overline{y}$) and performance decay ($\Delta y$) are indicated. $\pm\sigma$ is plotted as transparent.} \label{plot_fscil}
\end{figure*}

\noindent\textbf{Results.} The left plot of Figure~\ref{plot_fscil} shows the \emph{traditional} FSCIL setting. In this setting, the usual metric for comparison is performance decay $(\Delta y)$, a difference between first and last session incremental accuracy. This metric is consistent with the goal of few-shot class-incremental learning to prevent catastrophic forgetting; however, the results are not entirely clear when the lowest performance decay (best case) has a lower incremental accuracy (worst case) over a majority of the test. Attempts have been made to minimize this ambiguity by re-using the first session training or selecting hyper-parameters such that the first session has similar accuracy to baseline methods~\citep{tao2020fscil}. However, hyperparameters and full model architecture are not always shared, such that a major disadvantage of the setting is reproducibility~\citep{gibneycould}; moreover, it is unrealistic for a robot to train on half of the dataset before entering a new environment. Regardless, in \emph{traditional} FSCIL, the best performer in performance decay is CBCL ($-17.6\%$), an improvement of $2.4\%$ over FACT. The newer CBCL-PR ranks third in terms of performance decay ($-22.2\%$). It should be noted that other works have addressed more naturalistic learning paradigms, as opposed to the traditional FSCIL setting. For example Ren et. al define a new learning setting that is not based on episodes of training and testing but rather online, continual learning~\citep{ren2020wandering}.  

The right plot of Figure~\ref{plot_fscil} shows the \emph{pure} FSCIL setting. The results are less dependent on the accuracy of the first session, which makes for a fairer overall comparison; furthermore, this setting is more realistic for robots learning in unknown environments. The primary metric for comparison in this setting was average incremental accuracy $(\overline{y})$, which naturally considers performance decay $(\Delta y)$ along with the rate of decay. The best performer in this setting was CBCL-PR ($48.4\%$), an improvement of $2.3\%$ over CEC.  

\subsection{CBCL-PR Algorithm Details}
CBCL-PR is an updated version of CBCL~\citep{ayub2020cognitively, Ayub_BMVC20} and it is composed of the following primary components: fixed feature extractor, agg-Var clustering, and the generation of pseudo-exemplars for test prediction. 

 In each increment, the learner receives the training examples (images) for new classes. Feature vectors of the images are generated using a pre-trained CNN \emph{feature extractor}. With the exception of the preliminary study, the feature extractor was a Resnet-34 model pre-trained with ImageNet. 
 
 The learner applies \emph{Agg-Var clustering} on the feature vectors of new classes. Agg-Var clustering, inspired by the concept learning models of the hippocampus and the neocortex \citep{zeithamova12,Mack18}, enables the learner to discriminate between classes and consolidate these classes into long-term memory. Within the cluster space, each new class is initialized by creating a centroid of a new cluster using the first feature vector in the training set. Next, each additional feature vector $x_i^j$ (i.e., $i$-th image in class $j$) is compared to all the existing centroids for class $j$. If the Euclidean distance between $x_i^j$ and the closest centroid is greater than a pre-defined distance threshold $D$, a new centroid is created for class $j$ and equated to $x_i^j$. If the distance is less than $D$, the closest centroid is updated with a weighted mean: the $n$-th update ($n>1$) is calculated by equation (\ref{eq:mean}) below: 

\begin{equation}
    \begin{aligned}
        n \bar{x}_{n} &= x_n + (n-1)\bar{x}_{n-1}
        \label{eq:mean}
    \end{aligned}
\end{equation}
 
\noindent Note that prior training data is not needed to calculate a new centroid, as the old centroid $\bar{x}_{n-1}$ and new feature vector $x_n=x_i^j$ are sufficient. This process results in a collection of centroids for the class $j$, $C^j = \{c_1^j, ..., c_{N_j}^j\}$, where $N_j$ is the number of centroids for class $j$. 

\begin{figure}
\begin{center}
\includegraphics[width=0.8\textwidth]{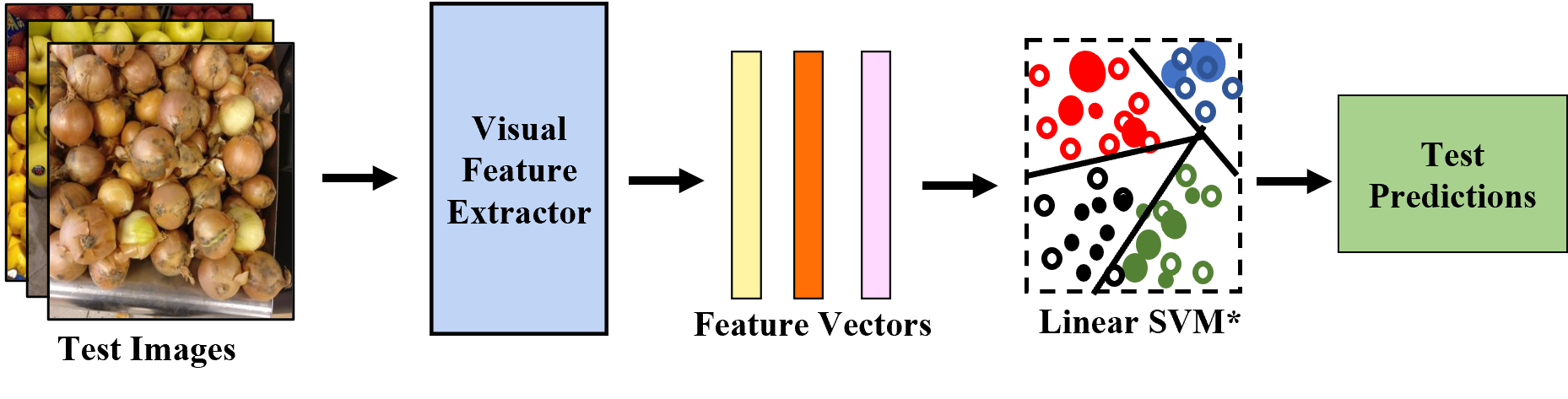}
\end{center}
\caption{This flow summarizes the testing phase of FIASco, where extracted features from test images are passed through the trained linear SVM to make predictions.} \label{testingPhase}
\end{figure}

In the update of the cluster space, covariance matrices of new clusters are recorded prior to the discarding of training data. These covariance matrices are used to generate a Gaussian distribution of \emph{pseudo-exemplars} centered on their respective centroid. A linear SVM\footnote{For slightly higher accuracy, a shallow neural net was used in the preliminary study. Either classifier can be used.} is trained using pseudo-exemplars of the old classes and feature vectors of the new classes. During testing, feature vectors of the test images are generated using the pre-trained CNN feature extractor and passed through the linear SVM to classify test images based on their feature vectors. This process is shown in Figure~\ref{testingPhase}.

\subsection{A* Algorithm Details}
In the final experiment, the FIASco algorithm was demonstrated on a Pepper robot navigating an indoor environment. A* path planning was used in order to get from point A (current location) to point B (goal location). The A* algorithm~\citep{hart1968formal} is a common search method that incrementally extends the path until a goal state is reached (or until the maximum number of iterations have been attempted, i.e., failure). Specifically, the agent extends the path with the next node ${n}$ minimizing the cost function (equation \ref{eq:astar}), where $f(n)$ is the total cost, $h(n)$ is the path distance from start, and $g(n)$ is the estimated path distance to goal. 

\begin{equation}
    \begin{aligned}
        f(n) = h(n) + g(n)
        \label{eq:astar}
    \end{aligned}
\end{equation}

In order to use this algorithm, the robot first used range sensors to compute a two-dimensional grid map. With an RGB camera, the robot localized objects and placed them on this two-dimensional grid map. Based on which object was most desirable, a goal location was selected. The A* algorithm was used to determine a path, given current and goal locations, and the two-dimensional map of the relative surroundings.

\subsection{Computational Cost Details}
\begin{figure*}
\begin{center}
\includegraphics[width=0.9\textwidth]{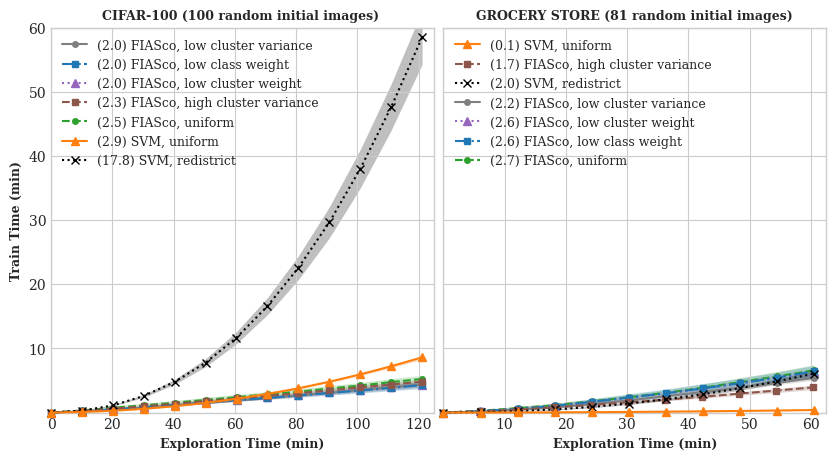}
\caption{Train time versus exploration time in Minecraft simulation. Note that average incremental train time is indicated and $\pm\sigma$ is plotted as transparent. }\label{cost}
\end{center}
\end{figure*}

The batch learner (SVM) and clustering approach (FIASco) both rely on a support vector machine with linear kernel to make test predictions. The data used to train the classifier is the same data type (thus, same dimension), so a comparison of computational complexity depends solely on the number of data points. For the batch learner, this collection of data includes every training instance the learner has collected to a given iteration. Conversely, the clustering approach of FIASco uses incoming data to update the centroids of the cluster space, for which the centroids (fewer than the total training instances) are the data points; however, the pseudo-exemplars should also be considered. Fortunately, the number of pseudo-exemplars are fixed per class and do not grow without bound. These observations can be seen in a plot of training time versus run time, shown in Figure~\ref{cost}. 

With the CIFAR-100 and Grocery store datasets, the FIASco only uses 5 and 40 pseudo-exemplars per class, respectively (see Section~\ref{Section:exp_malmo}). Additionally, the agent is permitted to explore longer in the CIFAR-100 environment, since the dataset is much larger. These factors explain the trend of computational cost. Initially, the difference between training instances and centroids is minimal, as the agent sees many new objects, creating many new centroids for a majority of incoming data. Furthermore, as the agent has not explored much of the environment, the total number of training points is small, as compared to the total number of pseudo-exemplars. Of course, as time progresses, the impact of the pseudo-exemplars is reduced relative to the total number of data points. Here, clustering would see much benefit in the long term.

\subsection{Tabular Results}
This section includes the tabular results from the simulation (Section~\ref{Section:exp_malmo}) and real-world (Section~\ref{Section:exp_pepper}) experiments.

\begin{table}[h!]
\caption{Test prediction accuracy (\%) as a function of run time, as the agent learns in Minecraft with CIFAR-100 data. This data corresponds to the left side of Fig.~\ref{plot_malmo}. The highest accuracy at a given increment is indicated with bold font.}\label{tab_malmo}
\centering
\begin{tabular}{l r r r r r r r r r r r}
\\ \textbf{Method} &  \textbf{0} & \textbf{12} & \textbf{24} & \textbf{36} & \textbf{48} & \textbf{60} & \textbf{72} & \textbf{84} & \textbf{96} & \textbf{108} & \textbf{120 min.}\\
\hline
FIASco, low cluster weight & 17.9 & 29.5 & 38.4 & 42.5 & 44.8 & 46.3 & 47.3 & 47.9 & 48.4 & 49.0 & 49.4\\
FIASco, high cluster variance & 17.8 & 23.9 & 27.9 & 29.6 & 30.7 & 32.3 & 33.9 & 35.3 & 35.9 & 36.4 & 36.8\\
FIASco, low cluster variance & 17.9 & \textbf{29.9} & 38.0 & 41.8 & 43.0 & 43.8 & 44.5 & 45.0 & 45.3 & 45.7 & 45.9\\
FIASco, low class weight & 17.9 & 29.0 & \textbf{40.8} & \textbf{44.5} & \textbf{46.9} & \textbf{47.7} & \textbf{48.8} & \textbf{49.8} & \textbf{50.4} & \textbf{50.7} & \textbf{51.2}\\
FIASco, uniform & 17.9 & 27.7 & 34.8 & 39.3 & 42.7 & 45.3 & 46.8 & 48.0 & 49.0 & 49.9 & 50.6\\
SVM, redistrict & 17.8 & 25.7 & 32.5 & 37.4 & 40.0 & 43.2 & 45.2 & 46.8 & 48.1 & 49.4 & 50.4\\
SVM, uniform & 17.8 & 27.0 & 34.2 & 38.0 & 41.1 & 43.3 & 45.6 & 46.6 & 47.6 & 48.8 & 49.8\\
CEC, uniform & 8.2 & 9.4 & 10.3 & 10.6 & 11.4 & 11.8 & 12.0 & 12.1 & 12.6 & 12.6 & 12.6\\
FACT, uniform & 8.3 & 10.5 & 11.4 & 11.8 & 12.3 & 12.5 & 12.7 & 13.0 & 13.2 & 13.2 & 13.3\\
\end{tabular}
\end{table}

\begin{table}
\caption{Test prediction accuracy (\%) as a function of run time, as the agent learns in Minecraft with Grocery store data. This data corresponds to the right side of Fig.~\ref{plot_malmo}. The highest accuracy at any point is indicated with bold font.}\label{tab_malmo}
\centering
\begin{tabular}{l r r r r r r r r r r r}
\\ \textbf{Method} &  \textbf{0} & \textbf{6} & \textbf{12} & \textbf{18} & \textbf{24} & \textbf{30} & \textbf{36} & \textbf{42} & \textbf{48} & \textbf{54} & \textbf{60 min.}\\
\hline
FIASco, low cluster weight & 30.6 & 40.4 & 52.9 & 59.3 & 64.6 & 67.6 & 70.2 & 72.3 & 73.9 & 75.4 & 76.5\\
FIASco, high cluster variance & 30.6 & 36.5 & 41.3 & 44.0 & 47.2 & 49.2 & 49.9 & 51.8 & 53.0 & 53.9 & 54.7\\
FIASco, low cluster variance & 30.6 & 39.9 & 52.6 & 60.2 & 64.9 & 68.1 & 69.9 & 71.6 & 72.3 & 73.0 & 74.0\\
FIASco, low class weight & 30.6 & 39.6 & 52.1 & \textbf{60.4} & \textbf{66.8} & \textbf{69.3} & \textbf{71.1} & \textbf{73.4} & \textbf{75.0} & \textbf{75.9} & \textbf{76.2}\\
FIASco, uniform & 30.6 & 38.2 & 45.8 & 52.5 & 57.3 & 60.7 & 64.4 & 67.2 & 69.4 & 71.6 & 74.4\\
SVM, redistrict & 30.5 & 37.3 & 44.6 & 52.2 & 57.3 & 62.0 & 64.7 & 68.0 & 70.0 & 72.2 & 74.0\\
SVM, uniform & 30.5 & 37.4 & 44.9 & 50.8 & 56.7 & 61.6 & 65.4 & 68.7 & 71.3 & 73.0 & 74.8\\
CEC, uniform & \textbf{52.6} & 47.6 & 51.7 & 56.0 & 60.2 & 62.0 & 63.6 & 65.5 & 66.3 & 67.4 & 69.0\\
FACT, uniform & 51.4 & \textbf{50.3} & \textbf{54.0} & 57.9 & 61.1 & 62.4 & 63.9 & 66.2 & 66.7 & 67.2 & 68.0\\
\end{tabular}
\end{table}

\begin{table}
\caption{Test prediction accuracy (\%) as a function of requested classes, as Pepper learns with Grocery store data. This data corresponds to Fig.~\ref{plot_pepepr}. The highest accuracy at any point is indicated with bold font.}\label{tab_malmo}
\centering
\begin{tabular}{l r r r r r r r r r r r}
\\ \textbf{Method} &  \textbf{0} & \textbf{6} & \textbf{12} & \textbf{18} & \textbf{24} & \textbf{30} & \textbf{36} & \textbf{42} & \textbf{48} & \textbf{54} & \textbf{60 requests}\\
\hline
FIASco, low class weight & 54.5 & \textbf{57.2} & \textbf{60.2} & \textbf{58.2} & \textbf{60.2} & \textbf{60.9} & 61.5 & 63.9 & 62.2 & 66.6 & 67.9\\
FIASco, high cluster variance & 54.5 & 56.5 & 56.9 & 57.2 & 59.2 & 58.9 & \textbf{62.5} & 64.2 & \textbf{67.6} & \textbf{67.2} & \textbf{68.2}\\
FIASco, low cluster variance & 54.5 & 54.2 & 54.5 & 55.5 & 57.5 & 58.5 & 59.9 & 61.5 & 61.5 & 63.5 & 63.2\\
FIASco, low cluster weight & 54.5 & 52.2 & 52.8 & 56.9 & 59.9 & \textbf{60.9} & \textbf{62.5} & 61.9 & 62.9 & 62.5 & 63.2\\
FIASco, uniform & 54.5 & 54.8 & 55.2 & 56.2 & 59.2 & 59.9 & 60.9 & 63.9 & 65.2 & 66.2 & 66.6\\
SVM, uniform & 54.8 & 51.5 & 57.2 & 57.9 & 59.5 & 60.5 & 62.2 & \textbf{64.9} & 65.6 & 66.9 & 67.2\\
\end{tabular}
\end{table}

\subsection{Potential Field Details}

Regarding the forces used in the potential field, ACS methods were used to prioritize the order the classes. Then, the classes were divide by quartile. The top 25 percent of classes received the highest magnitude of attraction, the next 25 percent received the next highest attraction and so on. Table~\ref{magTable} describes the different splits attempted. The actual splits used in the simulation experiment were those described as \textit{mod 1}. Note that $-$ and $+$ reflect attraction and repulsion, respectively, while the integer that follows reflects magnitude.

\begin{table}[h!]
\caption{A breakdown of attraction splits per quartile of attractive classes.}\label{magTable}
\centering
\begin{small}
\begin{tabular}{l r r r r r}
\\ \textbf{Magnitude Descriptor} &  \textbf{Q1} & \textbf{Q2} & \textbf{Q3} & \textbf{Q4}\\
\hline
attract $+$ repulse &  $ -20 $ & $ -10 $ & $ +10 $ & $ +20 $\\
attract $+$ ignore &  $ -20 $ & $ -10 $ & $  00 $ & $  00 $\\
mod 1 &  $ -20 $ & $ -10 $ & $ +05 $ & $ +05 $\\
mod 2 &  $ -20 $ & $ -10 $ & $ -05 $ & $ -05 $\\
mod 3 &  $ -20 $ & $ -10 $ & $ -10 $ & $ -10 $\\
attract only &  $ -20 $ & $ -20 $ & $ -20 $ & $ -20 $\\
\end{tabular}
\end{small}
\end{table}

\subsection{Dataset Notes}
\noindent\textbf{Reason for datasets.} The CIFAR-100~\citep{krizhevsky2009learning} dataset was chosen as a common benchmark in continual learning tasks~\citep{Rebuffi_2017_CVPR, lopez2017gradient}. The Grocery Store dataset~\citep{klasson2019hierarchical} represents an ecologically valid dataset that is very close to a real-world dataset. The authors specify the data was collected from 18 different grocery stores with realistic features such as misplaced objects, varying distances, angles, lighting conditions, etc.

\noindent\textbf{Nature of training data.} In both experiments, the data was modified to have a 90:10 stratified train-test split. The test data (10 percent) was used for the evaluation of the classifier at every iteration. In the simulation experiment, the learner selects from Minecraft objects within an observable distance. Each Minecraft object has a 1:1 mapping to classes in the offline dataset, either CIFAR-100 or Grocery Store. Thus, when the agent is near $n$ Minecraft objects, it processes as being near $n$ specific classes and selects a class. The learner receives training instances from the offline dataset corresponding to the selected class.  In the real-world experiment, the learner detects real objects using an RGB camera, predicts the classes for which the objects belong with its classifier, and then selects from the predicted classes to train. The learner receives training instances from the offline dataset corresponding to the real object requested (from pointing). When the learner receives training instances from the offline dataset, it either stores the training features (SVM learner) or uses the training features to update the cluster space and ACS methods (FIASco). 

\noindent\textbf{Future work.} Evaluating these methods in a more cluttered or less structured environment could also be an interesting problem. Testing on such cluttered data for ACS has not been explored in most prior works~\citep{lomasky2007active, wu2011active, kottke2021probabilistic}. This application might also require object segmentation or object detection, which is currently out of the scope of this work. We do note that grocery stores are highly structured environments in which products are naturally categorized and organized in a logical manner. Moreover, items are placed on the shelves in a reasonably uncluttered manner. 

\end{document}